\documentclass[]{article}
\usepackage[letterpaper]{geometry}
\usepackage{amta2022}
\usepackage{times}
\usepackage{url}
\usepackage{latexsym}
\usepackage{natbib}
\usepackage{layout}

\usepackage[T1]{fontenc}

\usepackage[utf8]{inputenc}
\usepackage{microtype}
\usepackage{booktabs}
\usepackage{graphicx}
\usepackage{multirow}
\usepackage{enumitem}
\usepackage{amssymb}
\usepackage{xcolor}
\usepackage{tikz}
\usetikzlibrary{shapes}
\usepackage{CJKutf8}

\newcommand\meta{$^\heartsuit$}
\newcommand\unc{$^\spadesuit$}
\newcommand\msr{$^\clubsuit$}

\begin{document}

\title{How Robust is Neural Machine Translation to Language Imbalance in Multilingual Tokenizer Training?}

\author{\centering \textbf{Shiyue Zhang\unc\thanks{\ \ Work done during an internship at Meta AI.}, Vishrav Chaudhary\msr\thanks{\ \ Work done while at Meta AI.}, Naman Goyal\meta}, \\ \textbf{James Cross\meta, Guillaume Wenzek\meta, Mohit Bansal\unc, and Francisco Guzmán\meta} \\
  \unc UNC Chapel Hill $\;\;$ \meta Meta AI $\;\;$ \msr Microsoft Turing \\
  {\tt \{shiyue, mbansal\}@cs.unc.edu} $\;\;$ {\tt vchaudhary@microsoft.com} \\
  {\tt \{naman, jcross, guw, fguzman\}@fb.com} \\
}

\maketitle
\pagestyle{empty}

\begin{abstract}
  A multilingual tokenizer is a fundamental component of multilingual neural machine translation. It is trained from a multilingual corpus. Since a skewed data distribution is considered to be harmful, a sampling strategy is usually used to balance languages in the corpus. However, few works have systematically answered how language imbalance in tokenizer training affects downstream performance. In this work, we analyze how translation performance changes as the data ratios among languages vary in the tokenizer training corpus. We find that while relatively better performance is often observed when languages are more equally sampled, the downstream performance is more robust to language imbalance than we usually expected. Two features, \emph{UNK rate} and \emph{closeness to the character level}, can warn of poor downstream performance before performing the task. We also distinguish language sampling for tokenizer training from sampling for model training and show that the model is more sensitive to the latter. 
\end{abstract}

\section{Introduction}
\label{sec:intro}
Tokenization is an essential pre-processing step for most natural language processing (NLP) models. Out of different tokenization methods, subword tokenization \citep{schuster2012japanese, sennrich2016neural, kudo2018subword} has become \emph{de facto}. 
The creation of each subword is mainly based on frequency, i.e., if two characters often appear together, they will be merged into a subword. 
When more than one language is involved, instead of learning independent tokenizers for each language, people usually train a joint tokenizer from a multilingual training corpus \citep{sennrich2016neural, devlin2019bert}. In this case, the data percentage of each language directly affects how it will be represented. If one language dominates the training corpus, its words will mostly stay intact and hardly be split into subwords. In contrast, if the language gets starved, it will be excessively tokenized into characters, thus, the sentence length will be dramatically longer, and some tokens will be considered as unknown (UNK). Moreover, Neural machine translation (NMT) is known to be bad at dealing with long sentences and UNKs \citep{koehn2017six}. 

Recently, there is an increasing interest in building multilingual neural models that can process multiple languages \citep{devlin2019bert, liu2020multilingual, xue2021mt5}. A challenge that comes with this important task is to balance languages with different amounts of training data to avoid low-resource languages being under-represented, e.g., being excessively tokenized and being less seen by the neural models.
Existing works usually adopt the \emph{temperature sampling} strategy \citep{devlin2019bert, arivazhagan2019massively, conneau2019cross, xue2021mt5} (see detailed descriptions in Section~\ref{sec:related_multi}). However,
very few investigations of how language imbalance affects downstream performance have been conducted. Additionally, whenever previous works apply a certain language balancing strategy, they apply it for both \emph{tokenizer training} (balancing the data sizes of different languages in the tokenizer training corpus) and \emph{model training} (balancing the frequencies of sampling training mini-batches from different languages). Until now, it is unclear how each of them separately affects the downstream performance. 

In this work, we specifically investigate how robust NMT is to language imbalance in tokenizer training. We propose to vary the data ratio among languages in the tokenizer training corpus while keeping other settings (e.g., language sampling for model training, hyperparameters) fixed, and then check how translation results change (Section~\ref{sec:bi-exp}). 
However, finding the best data ratio through performing the downstream task is highly expensive. To provide an easy indication of tokenizer quality (or early prediction of downstream performance), we examine two intermediate features (Section~\ref{sec:bi-feature}): \emph{UNK rate} -- the average percentage of unknown words (marked with the UNK token) in each sentence, and \emph{closeness to the character level} -- the average sentence length in subwords divided by sentence length in characters.  

Through comprehensive bilingual and multilingual experiments among 8 languages (English, Tagalog, Icelandic, Danish, Indonesian, Tamil, Greek, and Chinese), we make the following \textbf{five main observations}: (1) NMT performance is more robust to language imbalance than we usually expected: especially when languages share scripts, performance drops only happen when the data ratio of two languages is as disparate as 1:10$^5$. (2) Better performance is often achieved when languages are more balanced: we observe moderate Pearson correlations between translation performance and the degree of language balance. (3) English can ``never'' be starved because English tokens often appear in the ``monolingual'' data of other languages. 
(4) In most cases, the two features (UNK rate and closeness to the character level) can hint at poor translation performances before performing the task. 
(5) NMT is more sensitive to language imbalance in model training than in tokenizer training. See more observations and discussions in Section~\ref{sec:bi} and Section~\ref{sec:multi}. 

Based on these observations, we provide the following \textbf{two practical suggestions}: 
(1) Instead of using temperature sampling, we want to keep the involved languages as balanced as possible when training a new multilingual tokenizer; (2) Before applying a pretrained tokenizer for new experiments or languages, we suggest evaluating it on a development set to make sure every language's UNK rate is low (lower than around 3.7\%, according to our experiments) and every language's closeness to the character level is also low (lower than around 0.87, according to our experiments).\footnote{\label{footnote:1}The exact threshold numbers (3.7\% and 0.87) are based our experiments and may not always hold. But we believe that the concept of checking the two features (UNK rate and the closeness to the character level) to make sure they are low enough should generalize to other situations.}

\section{Related Works}
\label{sec:related}
\subsection{Tokenization Methods}
Over the years, many tokenization methods have been proposed. Early works tokenize texts into ``words'', e.g., \texttt{MosesTokenizer} \citep{koehn2007moses}. However, language-specific tokenizers are needed and it often ends up with many rare tokens or UNKs. 
\emph{Subword tokenization} methods were introduced to tackle this problem: the idea is to keep frequent words intact and split rare words into frequent subwords. 
Subword tokenization has become \textit{de facto}. \citet{schuster2012japanese} introduce \texttt{WordPiece} that starts from all characters and gradually merges two units that improve language model (LM) likelihood the most. \citet{sennrich2016neural} propose to learn subwords via Byte-Pair Encoding (\texttt{BPE}) that merges the most frequent pairs first. 
\citet{kudo2018subword} propose a \texttt{unigram LM} method. It starts with a large vocabulary and gradually prunes it down to the desired size by removing tokens that are less likely to reduce the unigram LM likelihood. 
Subword tokenization methods usually assume the existence of pre-tokenization (e.g., split by whitespaces), which can cause de-tokenization ambiguity. To address this, \texttt{SentencePiece} \citep{kudo2018sentencepiece} treats whitespace as a special symbol, \_ (U+2581), to achieve \textit{lossless} tokenization. This toolkit supports both BPE and unigram LM tokenization. 
Despite the success of subword tokenization, it is no panacea, e.g., it is out-of-the-box and agnostic to the downstream tasks, it has no guarantee that subwords are meaningful, and it is vulnerable to typos \citep{sun2020adv}. Thus, ``tokenization-free'' models that directly encode characters or bytes or visuals have been introduced \citep{chung2016character, lee2017fully, salesky-etal-2021-robust} and are gaining more interest recently \citep{clark2021canine, xue2021byt5, tay2021charformer}. 

\subsection{Multilingual Tokenization}
\label{sec:related_multi}
Along with the development of multilingual models, people start to deal with multilingual tokenization. \citet{firat2016multi} learn a 30K subword vocabulary for each language. \citet{johnson2017google} oversample languages to the same size and train a joint WordPiece vocabulary. Recent multilingual works adopt this joint-vocabulary method, but instead of oversampling languages to the same size, they use \emph{temperature sampling} which was first introduced by multilingual BERT (mBERT) \citep{devlin2019bert}. Given the original data distribution $\{p_i\}_{i=1}^N$, where $p_i$ is the percentage of the $i^{th}$ language out of the total N languages, they exponentiate each $p_i$ by a factor $S$ ($0\leq S \leq1$), i.e., $p_i^S$. Then, they re-normalize them to get the new percentage of each language $\hat{p}_i = p_i^S/\sum_ip_i^S$, and they sample data according to the new percentages. Essentially, it down-samples high-resource languages and up-samples low-resource ones. \citet{arivazhagan2019massively} redefine $S$ as $\frac{1}{T}$ ($T$ stands for temperature). $S$ is usually set around 0.2 to 0.7, i.e., \emph{flattening the data distribution to some degree but not to uniform distribution} \citep{arivazhagan2019massively, conneau2019cross, conneau2020unsupervised, xue2021mt5}. \citet{chung2020improving} challenge this joint vocabulary recipe and propose to learn separate vocabularies for each language cluster. 

\subsection{Analysis and Assessment of Tokenization}
Since the choice of tokenization algorithm and training parameters affects downstream performances, previous works try to analyze or assess tokenization. Some works focus on the choice of vocabulary size. \citet{gowda-may-2020-finding} show that the near-optimal vocabulary size is when 95\% of tokens appear more than 100 times in the training set. \citet{ding-etal-2019-call} find that low-resource language pairs usually require fewer than 4K BPE merge-operations.  \citet{volt} evaluate vocabularies by Marginal Utility of Vocabularization and propose to tokenize as well as find the optimal vocabulary size via the Optimal Transport method. Some other works compare different tokenization algorithms. \citet{domingo2018much} compare 5 tokenizers and the best tokenizer varies across language pairs. \citet{bostrom2020byte} compare BPE to unigram LM for LM pretraining and show that unigram LM learns subwords that align better with morphology and leads to better performance. 

When multiple languages are involved, \citet{gerz2018relation} show that language typology is correlated with LM performance. \citet{judit2019exploring} find that mBERT \citep{devlin2019bert} vocabulary are dominated by subwords of European languages, and the tokenizer keeps English mostly intact while generating different distributions for morphologically rich languages. \citet{rust2020good} observe that mBERT usually performs worse than its monolingual counterparts because language-specific tokenizers keep the language from being excessively tokenized. Some works compare different \emph{temperature sampling} factors (S or T). \citet{arivazhagan2019massively} compare multilingual translation results of using temperature T=1, 5, 100, and find that T=5 works best. \citet{xue2021mt5} compare multilingual LM performances for sampling factor $S$=0.2-0.7 and find that $S=0.3$ is the best. However, note that the performance difference is a joint effect of both tokenizer and model training because the sampling is applied for both. Differently, in this paper, we analyze how language imbalance specifically in multilingual tokenizer training affects the downstream translation performance.
\section{Bilingual Experiments}
\label{sec:bi}
To examine how language imbalance in tokenizer training affects downstream translation performance, we first conduct English-centric bilingual experiments in which imbalance only happens for one single pair of languages (i.e., English and anther language). This gives us a more controlled setting compared to when multiple languages are involved. Nonetheless, we conduct multilingual experiments in Section~\ref{sec:multi}. Our main methodology is to keep the total tokenizer training data size fixed, gradually ``starve'' English, i.e., reduce English data percentage and increase the percentage of the other language, and then check the downstream translation performance. It is important to note that, to separate the influences of tokenizer and model, we use different data for tokenizer training and model training, and the model training data are always the same. 

\subsection{Experimental Setup}
\label{sec:bi-exp}
\paragraph{Languages.}  We experiment with 8 languages: English (en), Tagalog (tl), Icelandic (is), Danish (da), Indonesian (id),  Tamil (ta), Greek (el), Chinese (zh). The data statistics are shown in Table~\ref{tab:lang}. According to \textsc{Flores}101 \citep{DBLP:journals/corr/abs-2106-03193}, Icelandic, Tamil, and Tagalog are \textit{low-resource} ($\leq$ 1M bitext), while Danish, Greek, Chinese, and Indonesian are \textit{mid-resource} ($\leq$ 100M bitext). Tagalog, Icelandic, Danish, and Indonesian are Latin languages and thus share scripts with English; while Tamil, Greek, and Chinese are non-Latin. 

\paragraph{Translations.} We conduct English-centric bilingual translations in 14 directions: en-tl, tl-en, en-is, is-en, en-da, da-en, en-id, id-en, en-ta, ta-en, en-el, el-en, en-zh, zh-en. We train one translation model for each direction.

\begin{table}
\begin{center}
\resizebox{0.5\textwidth}{!}{%
\begin{tabular}{lllll}
\toprule 
\textbf{Language} & \textbf{Code} & \textbf{Script} & \textbf{En-* bitext} & \textbf{Mono. text} \\
\midrule
English & en & Latin & - & 2B \\
Tagalog & tl & Latin & 71K & 107M \\
Icelandic & is & Latin & 1M & 37M \\
Danish & da & Latin & 11M & 343M \\
Indonesian & id & Latin & 39M & 1B \\
Tamil & ta & Tamil & 97K & 68M  \\
Greek & el & Greek & 24M & 200M\\
Chinese & zh & Han & 38M & 293M \\
\toprule
\end{tabular}
}
\end{center}
\vspace{-10pt}
\caption{8 languages in our experiments. K/M/B stands for thousand/million/billion. Mono. stands for monolingual. Numbers are the number of sentences (pairs).}
\vspace{-7pt}
\label{tab:lang}
\end{table}

\paragraph{Variables.} For each translation direction, we have the following controlled, independent, and dependent variables:

\noindent \emph{Controlled variables}
\begin{itemize}
    \vspace{-4pt}
    \item Tokenizer training data: We use the same monolingual data as \textsc{Flores}101 \citep{DBLP:journals/corr/abs-2106-03193}. The total monolingual data sizes of each language are listed in Table~\ref{tab:lang}. We sample from these monolingual datasets to get the desired tokenizer training data size.\footnote{To minimize sampling influence, we shuffle each monolingual dataset once and then always sample the first X sentences.} We keep the total tokenizer training data size as 2M, which contains $x$\% English data and $1-x$\% data of the other language.
    \vspace{-4pt}
    \item Tokenizer parameters: We use SentencePiece model (SPM) with unigram LM algorithm \citep{kudo2018subword, kudo2018sentencepiece}. We set vocabulary size as 5K,\footnote{We set vocabulary size as 5K because (1) a small vocab size makes the ``competition'' between languages more ``fierce'' and thus makes it easier to show the problem of language imbalance, and (2) it resembles a multilingual setting: \textsc{Flores}101 uses a 256K vocabulary for 101 languages -- 2.5K tokens per language on average.} total training data size as 2M, and character coverage as 0.99995 (or 0.995 when Chinese is involved because Chinese has a richer character set).
    \vspace{-4pt}
    \item Translation training data: We also use the same parallel data as \textsc{Flores}101 \citep{DBLP:journals/corr/abs-2106-03193} (data sizes are in Table~\ref{tab:lang}). As mentioned above, we do not change this model training data across different experiments. And following previous works (Section~\ref{sec:related_multi}), we always use temperature sampling with $S=0.2$ for model training.
    \vspace{-4pt}
    \item Translation evaluation data: We evaluate on \textsc{Flores}101 \citep{DBLP:journals/corr/abs-2106-03193} dev sets and report results on its devtest sets.
    \vspace{-4pt}
    \item Translation model: Transformer \citep{vaswani2017attention} with 12-layer encoder and 12-layer decoder (Transformer 12-12).
    \vspace{-4pt}
    \item Model training and testing hyper-parameters: Adam optimizer \citep{kingma2015adam}, learning rate = 0.001, and beam size = 5. See more implementation details in~\ref{app:implement}.
    
\end{itemize}
\noindent \emph{Independent variable}
\begin{itemize}
    \vspace{-4pt}
    \item English data percentage in 2M tokenizer training data\footnote{We choose to directly vary the data percentage rather than sampling temperature because it grants us the flexibility of making high-resource languages hypothetically low-resource and experimenting with extreme data ratios (100\%: 0\%). }: we experiment with 9 different percentages (0\%, 0.001\%, 0.1\%, 10\%, 50\%, 90\%, 99.9\%, 99.999\%, 100\%). E.g., if we conduct en-zh/zh-en translations with English percentage=0.001\%, there are 20 English sentences and 2M - 20 Chinese sentences in SPM tokenizer training data. Hence, for each translation direction, we have 9 experiments with 9 different vocabularies. 
    See examples of how sentences are tokenized at different English percentages in Table~\ref{tab:examples} of~\ref{app:examples}.
\end{itemize}
\noindent \emph{Dependent variable}
\begin{itemize}
    \vspace{-4pt}
    \item Translation performance: we evaluate it by sentence-piece BLEU (spBLEU) \citep{DBLP:journals/corr/abs-2106-03193}\footnote{Computing BLEU \citep{papineni2002bleu} requires a tokenizer. However, not all languages have language-specific tokenizers available. spBLEU \citep{DBLP:journals/corr/abs-2106-03193} unifies the evaluation across languages by first tokenizing languages via a 256K multilingual SPM and then computing BLEU.} and chrF \citep{popovic-2015-chrf}. Metrics are computed by SacreBLEU \citep{post-2018-call}.\footnote{\url{https://github.com/ngoyal2707/sacrebleu/tree/adding_spm_tokenized_bleu}} We report the 3-seed average for each experiment.
\end{itemize}
\vspace{-4pt}

\subsection{Intermediate Features}
\label{sec:bi-feature}
Previous works have shown that without training downstream models, some intermediate features can be good indicators of the tokenizer's quality \citep{gowda-may-2020-finding, chung2020improving, volt}.
In this work, as the English data percentage varies, either English or the other language will get starved -- sentence lengths will become longer and unknown words (UNKs) will appear. Hence, we examine the following two features:
\begin{itemize}
    \vspace{-4pt}
    \item \emph{Closeness to the character level}, defined as the average $\frac{sentence\ length\ in\ subwords}{sentence\ length\ in\ characters}$. Some languages may intrinsically have longer sentence lengths than others. To be comparable across languages, we normalize it by the upper bound -- sentence length in characters. 
    \vspace{-4pt}
    \item \emph{UNK rate}, which is defined as the average $\frac{number\ of\ UNKs}{sentence\ length\ in\ subwords}$. Note that when the UNK rate increases, long unknown tokens will not get split into subwords, and thus the sentence length will be shorter and the closeness to the character level will decrease. 
    \vspace{-5pt}
\end{itemize}

\begin{figure*}
    \centering
    \includegraphics[width=0.98\textwidth]{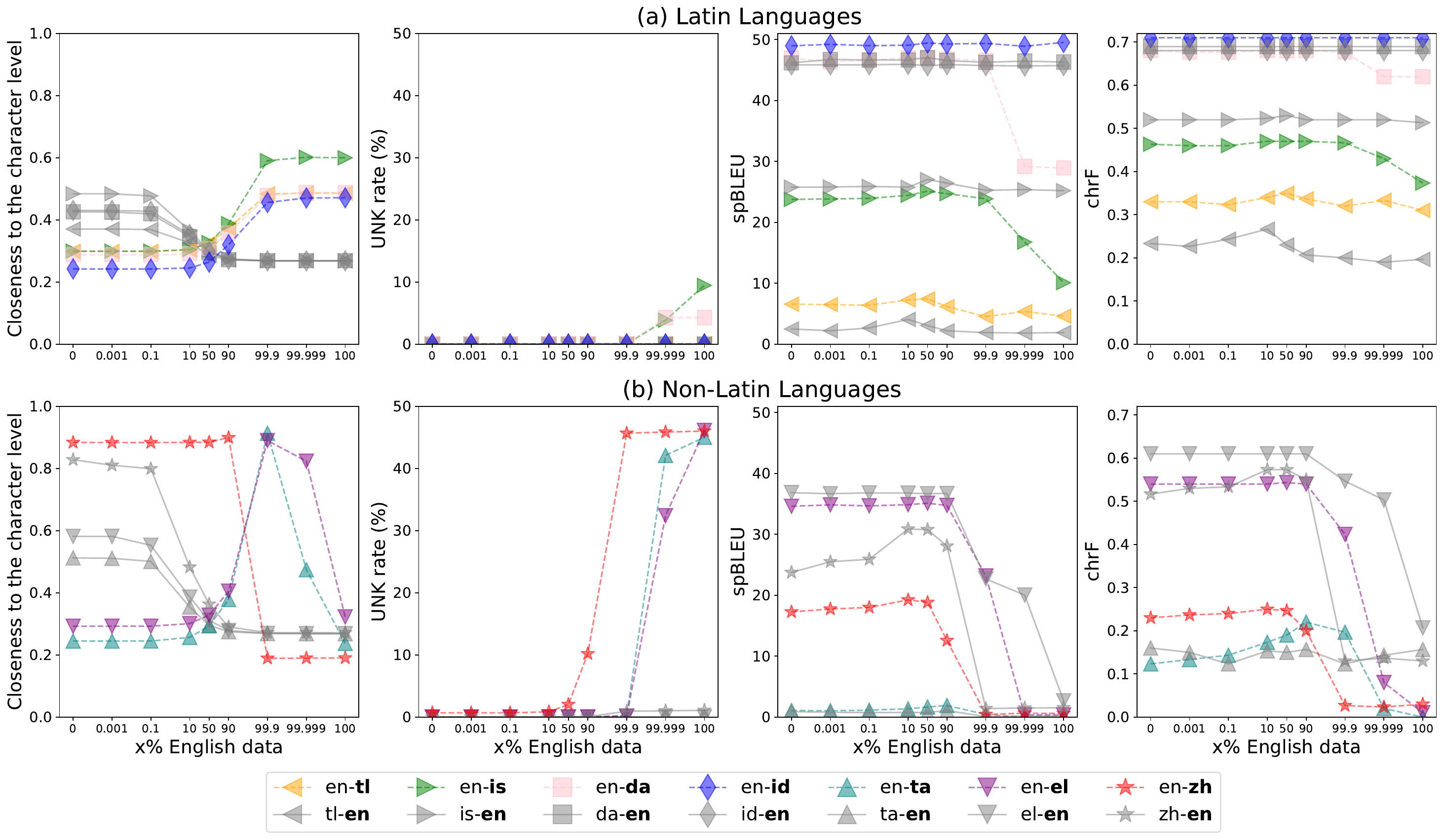}
    \vspace{-7pt}
    \caption{Results of our main bilingual experiments. 
    Marker shapes denote the language pairs; dash or solid lines represents out-of-English or into-English directions; colors are for each target language. E.g., {\color{teal}-}{\color{teal}-}$\begingroup\color{teal}\blacktriangle\endgroup${\color{teal}-}{\color{teal}-} (en-\textbf{ta}) denotes Tamil features (\emph{Closeness to the character level} or \emph{UNK rate}) or English to Tamil translation results (\emph{spBLEU} or \emph{chrF} scores); {\color{gray}-}$\begingroup\color{gray}\blacktriangle\endgroup${\color{gray}-} (ta-\textbf{en}) represents English features or Tamil to English translation results. X axes are in log10 scale.}
    \vspace{-7pt}
    \label{fig:main_res}
\end{figure*}

The first two columns of Figure~\ref{fig:main_res} illustrate how the intermediate features change as the English data percentage changes. The first row (a) shows features of the 4 Latin languages, while the second row (b) is those of the 3 non-Latin languages. Note that both features are computed on \textsc{Flores}101 \citep{DBLP:journals/corr/abs-2106-03193} devtest sets. 
\vspace{3pt}

\paragraph{Closeness to the character level.} In Figure~\ref{fig:main_res} (a), as the English percentage increases, the closeness to the character level of English (gray markers) decreases while that of other languages (makers with other colors) increases. It is because when the English percentage gets larger, the other language's tokens will become rarer and be excessively tokenized into subwords. Differently, in Figure~\ref{fig:main_res} (b), though the trend of English stays the same, the trend of other languages first increases close to 1.0 and then decreases because UNKs start to appear. Even when English occupies 100\%, Latin languages still have sentence lengths much shorter than the sentence length in characters because they share scripts with English. In contrast, each of the 3 non-Latin languages reaches close to the character level at a certain point. English never have very long sentence lengths.
\vspace{4pt}

\paragraph{UNK rate.} In Figure~\ref{fig:main_res} (a), most UNK rates are trivial (close to 0), except that Icelandic (is) and Danish (da) have non-trivial UNK rates when English percentage $\geq$ 99.999\%. In Figure~\ref{fig:main_res} (b), all three non-Latin languages have very high UNK rates after the English percentage increases to a certain point. For example, Chinese (zh) has a 45.7\% UNK rate at English=99.9\%, and it is when its closeness to the character level drops dramatically. English always has trivial UNK rates.

\subsection{Translation Results}
The second two columns of Figure~\ref{fig:main_res} shows how the translation results change as the English data percentage changes. The first row (a) shows spBLEU and chrF scores of the 4 Latin languages, while the second row (b) are those of the 3 non-Latin languages. We obtain the following takeaways.

\paragraph{NMT performance is quite robust to language imbalance especially when languages share scripts.} It can be observed from Figure~\ref{fig:main_res} (a) that the performance stays quite stable across all English percentages for Latin languages. Performance drops only happen for English to Icelandic (en-is) and English to Danish (en-da) at extremely high English percentages ($\geq$99.999\%), i.e., only 20 Icelandic or Danish sentences are in the 2M tokenizer training data. And it still does not affect the translation performances of is-en and da-en. Differently, in Figure~\ref{fig:main_res} (b), the performance is less stable for non-Latin languages, but drops still happen when the English percentage is $\geq$90\%. English to Chinese (en-zh) drops at English=90\%. English to Tamil (en-ta)\footnote{Note that at English=99.9\%, Tamil's chrF scores only drop slightly while its spBLEU scores drop more significantly (en-ta drops from 1.9 to 0.4 and ta-en drops from 1.1 to 0.1).} and English to Greek (en-el) both drop at English=99.9\%. Similarly, into-English directions are more stable and get worse later (at higher English percentages). Surprisingly, in both (a) and (b), the translation performance usually stays stable or drops less significantly as the English percentage decreases to 0\%. 

\paragraph{Better performance is often achieved when languages are more balanced.} Out of the 14 translation directions, 12 directions get the best spBLEU scores between English=10\% to English=90\%. We evaluate the Pearson correlation between spBLEU scores and \emph{data ratios} of two languages. The data ratio is 1 when English=50\%, and it is 0 when English=0\% or 100\%, i.e., the more balanced the two languages are, the higher the data ratio is. The average correlation across 14 directions is 0.38 (moderate correlation \citep{cohen1988}). Thus, we are more likely to get a good performance when languages are more equally sampled. 

\paragraph{English can ``never'' be starved.} Initially, we were expecting a symmetric trend, i.e., if the performance drops as the English percentage increases, it should also drop when the percentage decreases. However, as shown in Figure~\ref{fig:main_res}, for both Latin and non-Latin languages, the performance stays relatively stable as the English percentage decreases to 0\%. We suspect that other languages' monolingual data contains many English words. First, we find that about 3.6\% and 2.6\% characters in Tamil and Chinese monolingual data are English characters (a-zA-Z) respectively. Then, we remove all English characters from Tamil or Chinese monolingual data and re-conduct the experiments of English=0.001\%. English-Tamil/Tamil-English spBLEU scores reduce from 1.0/0.8 to 0.0/0.3. Similarly, English-Chinese/Chinese-English spBLEU scores drop from 17.7/25.5 to 0.2/0.1. Hence, the results support our hypothesis. 

\paragraph{Closeness to the character level and UNK rate can warn of poor downstream performance.} We find that the translation performance usually drops greatly when the two features surpass some thresholds. As shown in Figure~\ref{fig:main_res} (a), both English to Icelandic (en-is) and English to Danish (en-da) get noticeably worse at English=99.999\%, and it is exactly when Icelandic and Danish have non-trivial UNK rates (3.9\% for is and 4.3\% for da). Similarly, in Figure~\ref{fig:main_res} (b), English to Chinese (en-zh) deteriorates at English=90\% when Chinese UNK rate is 10.2\%. English to Tamil (en-ta) and English to Greek (en-el) both drop at English=99.9\% when they have trivial UNK rates but their closeness to the character level are 0.91 and 0.89 respectively. 
Additionally, we examine whether the same pattern can still be observed when getting the features on a different evaluation set. We get features from the dev set and a subset of our training set (5000 sentence pairs). Despite the slightly lower thresholds (3.7\% UNK rate and 0.87 closeness to the character level), the same trends are observed. See details in Appendix~\ref{app:fea_eval}. Hence, we suggest checking these two features on an evaluation set before performing the task. Poor translation performances are likely to be obtained when any language's UNK rate is larger than around 3.7\% or its closeness to the character level is larger than around 0.87.

\subsection{Ablations}
\label{sec:bi-ablation}
Here, we want to verify our takeaways under several different experimental settings.

\paragraph{Reducing the translation model size or using BPE does not affect the robustness to language imbalance.} Model capacity can affect its robustness. Hence, we replace our default Transformer 12-12 \citep{vaswani2017attention} model with a smaller model, Transformer 6-6 (6-layer encoder and 6-layer decoder). The intermediate features are the same as Figure~\ref{fig:main_res}, and the translation results are illustrated in Figure~\ref{fig:smaller_model}. It has exactly the same trends as for the larger model (Figure~\ref{fig:main_res}). 
In addition, we verify if our takeaways can generalize to a different tokenization algorithm, BPE \citep{sennrich2016neural}. Figure~\ref{fig:bpe} shows that BPE gets very similar performances to unigram LM across all translation pairs. The same trends are also observed as Figure~\ref{fig:main_res} but with slightly higher thresholds. See details in ~\ref{app:bpe}. 

\paragraph{Increasing the vocabulary size can improve the robustness when languages do not share scripts.} Our default vocabulary size is 5K because it simulates a multilingual setting (see footnote2). However, earlier works used a larger vocabulary for bilingual experiments \citep{firat2016multi}. Intuitively, a larger vocabulary can be more robust to language imbalance because it has a larger capacity to include more infrequent words. Hence, we test a 32K vocabulary, and results are shown in Figure~\ref{fig:larger_vocab}. Compared to Figure~\ref{fig:main_res}, it has two distinctions: (1) For non-Latin languages, performance drops happen later: English to Chinese drops at 99.9\% (instead of 90\%) when Chinese UNK rate is 7.8\%; English to Tamil and English to Greek both deteriorate greatly at 99.99\% (instead of 99.9\%) when Tamil and Greek UNK rates are 42.3\% and 32.5\% respectively; (2) Surprisingly, translations between English and Tagalog perform obviously worse when English$\geq$99.999\%, despite Tagalog's trivial UNK rate and short sentence length. Overall, increasing the vocabulary size improves the robustness to language imbalance for translations between English and non-Latin languages but not for that between English and Latin languages.

\paragraph{Applying byte-fallback does not improve the robustness.}
Here, we apply the ``byte-fallback'' feature of \texttt{SentencePiece} \citep{kudo2018sentencepiece} which uses 256 UTF-8 bytes to represent unknown characters and thus eliminates UNKs. Figure~\ref{fig:byte} illustrates the results. As expected, UNK rates are all 0, while closeness to the character level can be larger than 1 because one character can be represented by multiple bytes. For Latin languages, noticeable drops still only happen for Icelandic and Danish starting from 99.999\%, but differently, they have 0 UNK rates and not high closeness to the character level (0.65 and 0.53). Moreover, performance drops are surprisingly more dramatic compared to Figure~\ref{fig:main_res}. The performances of all 3 non-Latin languages get worse at the same percentages as Figure~\ref{fig:main_res}, and the drop is more significant for Greek to English while less significant for Chinese to English. Overall, applying byte-fallback does not improve the robustness reliably. 

\paragraph{When English=100\%, adding characters of the non-Latin language to the vocabulary can improve the performance.} When English occupies 100\% of the tokenizer's training data, the tokenizer only ``knows'' English. Other Latin languages share scripts with English, so it shows surprisingly good generalizability. However, for non-Latin languages, near all tokens are UNKs, and thus translation performances are very poor. We wonder how much the performance will increase by simply adding the characters of the non-Latin language to the vocabulary. We conduct this experiment for each of the 3 non-Latin languages, and the results are shown in Table~\ref{tab:add_char}. Compared to the original setting (100\%), adding characters (100\%+char) dramatically improves the performance except for ta-en.
Despite that, for Tamil or Greek, it works greatly worse than the best we can achieve when Tamil or Greek data involves in tokenizer training. But, for Chinese, it outperforms the best results probably because one Chinese character is usually one ``word''.

\section{Multilingual Experiments}
\label{sec:multi}
Here, we move to a more complex multilingual setting. 
Similarly, we want to understand how the data percentages of the involved languages affect their downstream translation performance.

\subsection{Experiment Setup \& Features}
We still experiment with the 8 languages and the 14 translation directions, as introduced in Section~\ref{sec:bi-exp}. Differently, we use one model (Transformer 12-12) to conduct all the 14 translations at the same time. As a result, the model capacity for each translation direction is dramatically reduced. Most of the \emph{controlled variables} stay the same as Section~\ref{sec:bi-exp}, except that we increase the vocabulary size to 20K (maintaining around 2.5K per language) and increase the total tokenizer training data size to 10M. Since here we have 8-language data to train the tokenizer, we can not use the old \emph{independent variable}. Instead, we propose to first choose one language and then vary its percentage (0.001\%, 0.1\%, 1\%, 12.5\%, 25\%, 90\%) while keeping the other 7 languages equally weighted. So, if the selected language's percentage is 12.5\%, all 8 languages are equally weighted. We only use 4 languages (Tamil, Chinese, Icelandic, and English) as our selected languages and change the percentage of each of them. The \emph{dependent variable} is the same as before -- translation performance (spBLEU/chrF) on \textsc{Flores}101 \citep{DBLP:journals/corr/abs-2106-03193} devtest sets. We also examine the two \emph{intermediate features}: closeness to the character level and UNK rate.

\subsection{Results \& Ablations}
\label{sec:multi-results}
\begin{figure*}
    \centering
    \includegraphics[width=0.98\textwidth]{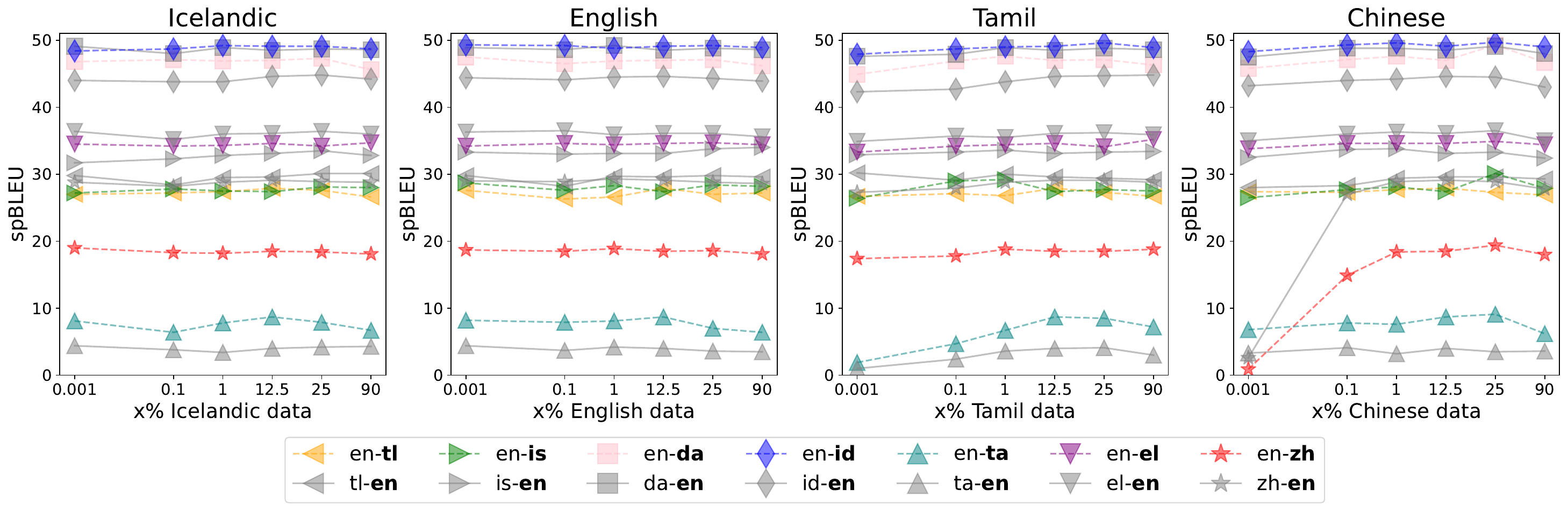}
    \vspace{-7pt}
    \caption{Translation results (spBLEU) of our main multilingual experiments. Marker shapes denote the language pairs (though all pairs share the same NMT model); dash or solid lines represents out-of-English or into-English directions; colors are for each language. E.g., {\color{teal}-}{\color{teal}-}$\begingroup\color{teal}\blacktriangle\endgroup${\color{teal}-}{\color{teal}-} (en-\textbf{ta}) denotes English to Tamil translation results; {\color{gray}-}$\begingroup\color{gray}\blacktriangle\endgroup${\color{gray}-} (ta-\textbf{en}) represents Tamil to English translation results. X axes are in log10 scale.}
    \vspace{-7pt}
    \label{fig:multi-spbleu}
\end{figure*}

Figure~\ref{fig:multi-spbleu} illustrates the translation performance evaluated by spBLEU (chrF in Figure~\ref{fig:multi-chrf} shares the same trends). Figure~\ref{fig:multi-feature} in~\ref{app:multi_main} shows the features.

\paragraph{NMT performance is still quite robust to language imbalance especially when languages share scripts.}
As shown in Figure~\ref{fig:multi-spbleu}, for the two Latin languages (Icelandic and English), varying their percentages almost does not affect the performances. It is expectable for English because it can ``never'' be starved. But Icelandic's performance drops at Icelandic=0.001\% (English=99.999\%) in bilingual experiments. We think it is because the involvement of multiple languages makes every language relatively less frequent, so the data ratio between Icelandic and any other language is not as disparate as 0.001:99.999 ($\approx$ 1:10$^5$). This is also reflected by the trivial UNKs of all languages in Figure~\ref{fig:multi-feature}. For the two non-Latin languages (Tamil and Chinese), first, varying their percentages affects their own performances greatly while the performances of other languages still stay stable. And, their own 
performances drop quickly below 12.5\% while dropping slower when percentages$\geq$12.5\%.

\paragraph{Better performance is also often observed when languages are more balanced.} In Figure~\ref{fig:multi-spbleu}, if we only consider the translation directions with great performance changes, i.e., Tamil and Chinese, they have relatively better performances around 12.5\% when languages are balanced. We define \emph{data ratio} as the lowest percentage of any language versus the highest percentage. So, the data ratio is 1 when the selected language's percentage is 12.5\%; while the data ratio is 0.07 when the selected language's percentage is 1\% ($\frac{0.01}{(1-0.01)/7}=0.07$). Then, we compute the correlation between spBLEU scores and data ratios for each of the 4 selected languages. The average correlation is 0.49 (moderate correlation \citep{cohen1988}), which is consistent with what we observe in bilingual experiments. 

\paragraph{Performance can drop without surpassing the thresholds of the two features.} For Chinese, a more obvious performance drop happens at 0.1\% following the indication of two features (UNK rate=5.4\% and closeness to the character level=0.97). However, for Tamil, though its performance drops at 1\%, it has a trivial UNK rate and not long sentence length. This is probably due to the greatly compressed model capacity for each language pair, compared to bilingual experiments. Hence, though surpassing the thresholds can often hint at poor performances, it is neither a sufficient nor necessary condition. 

\paragraph{Using byte-fallback still does not improve the robustness} We apply \emph{byte-fallback} under the setting of using Chinese as the selected language, and results are shown in Figure~\ref{fig:multi-zh-byte}. Compared to Figure~\ref{fig:multi-spbleu}, though we observe slightly more stable performance when Chinese$\geq$1\%, the translation result drops more dramatically when Chinese$\leq$0.1\%.

\paragraph{NMT is more sensitive to language imbalance in model training.} In both bilingual or multilingual settings, we find that the performance is quite robust to language imbalance and relatively better performance is often observed when languages are more balanced. In other words, we want to set sampling factor $S=0$, following the temperature sampling paradigm \citep{devlin2019bert}. However, many existing works show significantly different performances of different $S$, and the best $S$ is around 0.2 to 0.7 \citep{arivazhagan2019massively, conneau2019cross, xue2021mt5}. We think this inconsistency has resulted from the fact that we fix $S=0.2$ for model training while only varying it (via changing data percentages) for tokenizer training. We conjecture that NMT is more sensitive to language imbalance in model training. To verify this, first, we fix model training sampling $S=0.2$ and compare 3 tokenizer training sampling factors ($S=0, 0.3, 1.0$). Results are shown in the second row (starting with ``tokenizer'') in Table~\ref{tab:comparison}. Though with small differences (0.4, 0.1 points), $S=0$ overall works best. Second, we fix tokenizer training sampling $S=0$ and compare 3 model training sampling factors ($S=0, 0.2, 1.0$). As shown in Table~\ref{tab:comparison}, the differences are more prominent (1.4, 0.6 points), and $S=0.2$ overall works best. 
Hence, for tokenizer training, we want languages to be balanced, whereas, for model training, we want to flatten the original distribution to some degree but not to uniform distribution. And we want to pay more attention to sampling for model training because NMT is more sensitive to it. 

\begin{table*}
\begin{center}
\resizebox{0.99\textwidth}{!}{%
\begin{tabular}{ll|llllllll|llllllll|l}
\toprule 
& \multirow{2}{*}{\bf S} & \multicolumn{8}{c}{\bf *-en} & \multicolumn{8}{c}{\bf en-*} & \bf overall \\
\cmidrule(r){3-10}
\cmidrule(lr){11-18}
\cmidrule(l){19-19}
& & \bf tl &  \bf is & \bf da & \bf id & \bf ta & \bf el & \bf zh & \bf avg. & \bf tl & \bf is & \bf da & \bf id & \bf ta & \bf el & \bf zh & \bf avg.  & \bf avg. \\
\midrule
\multirow{3}{*}{tokenizer} & 0 &  29.6 & 33.1 & 48.5 & 44.6 & 4.0 & 36.1 & 29.1 & \bf 32.1 & 27.9 & 27.4 & 47.0 & 49.1 & 8.7 & 34.6 & 18.5 & 30.5 & \bf 31.3 \\
& 0.3 & 28.6 & 33.6 & 49.0 & 44.0 & 3.4 & 36.6 & 28.5 & 32.0 & 26.6 & 27.5 & 46.2 & 48.7 & 7.6 & 34.2 & 18.4 & 29.9 & 30.9 \\
& 1 & 29.0 & 32.4 & 48.4 & 44.1 & 3.4 & 35.6 & 28.8 & 31.7 & 27.5 & 29.0 & 47.8 & 49.7 & 7.6 & 34.6 & 19.1 & \bf 30.8 & 31.2 \\
\midrule
\multirow{3}{*}{model}& 0 & 28.2 & 32.6 & 47.4 & 41.6 & 3.6 & 34.2 & 26.7 & 30.6 & 26.9 & 27.8 & 45.9 & 47.3 & 6.9 & 33.1 & 17.1 & 28.3 & 29.9 \\
& 0.2 &  29.6 & 33.1 & 48.5 & 44.6 & 4.0 & 36.1 & 29.1 & 32.1 & 27.9 & 27.4 & 47.0 & 49.1 & 8.7 & 34.6 & 18.5 & \bf 30.5 & \bf 31.3\\
& 1 & 27.2 & 33.3 & 49.7 & 46.2 & 4.0 & 37.6 & 31.7 & \bf 32.9 & 16.9 & 25.7 & 47.8 & 50.1 & 3.4 & 35.7 & 19.8 & 28.5 & 30.7\\
\toprule
\end{tabular}
}
\end{center}
\vspace{-10pt}
\caption{Comparison of language sampling factors used in tokenizer or model training. All numbers are spBLEU. $S$ is the exponential factor used in temperature sampling (see Section~\ref{sec:related_multi}).}
\vspace{-7pt}
\label{tab:comparison}
\end{table*}

\section{Conclusion}
We systematically analyze how language imbalance in multilingual tokenizer training affects translation performances. Overall, we find that NMT performance is quite robust to language imbalance especially when languages share scripts. Better performance is often achieved when languages are more balanced. We suggest keeping the involved languages as balanced as possible in the tokenizer training corpus and evaluating pretrained tokenizers on an evaluation set to make sure no language's UNK rate $\geq$ around 3.7\% and no language's closeness to the character level $\geq$ around 0.87. We hope our work can provide some guidance for future multilingual tokenizer training and usage.

\section{Limitation}
This work is an empirical study. It is important to be aware that our observations and conclusions are made based on our experiments, which may or may not be generalizable to other settings. We try our best to include diverse languages, but still, our experiments are English-centric and at most have 8 languages involved. We tend to believe that the five main observations we made (as listed in the second last paragraph of Section~\ref{sec:intro}) are generalizable to other experimental settings. However, the exact thresholds of the two features (UNK rate and closeness to the character level) for indicating poor downstream performance may not be always hold (as mentioned in Footnote~\ref{footnote:1}). 

\section{Ethical Consideration}
The main ethical consideration of this work is that our experiments are many, so it is not very easy to finish them in a reasonable time without a decent number of computation resources. In the bilingual setting, we have 9 percentages, 14 directions, 5 ablations (including our basic setting), and 3 seeds. So we have 1890 experiments in total. Each experiment takes from less than 1 hour to about 2 days (based on the training data size) using 8 NVIDIA Tesla V100 Volta GPUs. In the multilingual setting, we only have 31 experiments in total, but each experiment takes 1.5 days using 64 GPUs. 

However, we expect that our empirical results can help guide the training and usage of multilingual tokenizers, so future works do not have to re-conduct these expensive investigations. Based on our results, the downstream performance is not highly sensitive to language imbalance in tokenizer training, and keeping languages as balanced as possible is a safe choice. Additionally, the two features (closeness to the character level and UNK rate) can serve as intermediate quality evaluators of pretrained tokenizers before performing the task.

\section*{Acknowledgments}
We thank the reviewers for their helpful comments and thank Angela Fan, Chau Tran, Xiang Zhou, Simeng Sun for helpful discussions. This work was done while SZ was interning at Meta AI and later supported at UNC by NSF-CAREER Award 1846185, ONR Grant N00014-18-1-2871, and a Bloomberg Data Science Ph.D. Fellowship. The views contained in this article are those of the authors and not of the funding agency. 

\small

\bibliographystyle{apalike}
\bibliography{amta2022}

\appendix

\section{Appendix}
\label{sec:appendix}

\subsection{Model Implementation Details}
\label{app:implement}
We implement translation models using fairseq.\footnote{\url{https://github.com/pytorch/fairseq}} During training, we use Adam optimizer \citep{kingma2015adam}, learning rate=0.001, and warmup for 2 epochs. We use batch size=4K tokens and gradient accumulation=4. For bilingual experiments, we use 8 NVIDIA Tesla V100 Volta GPUs for each experiment, and we run 3 seeds (2, 7, 42) for each experiment and report the average. For multilingual experiments, we use 64 GPUs and only run seed=2 for each experiment. We apply early stop with patience of 20 epochs. During testing, we use batch size=32 sentences and beam size=5.

\subsection{Compute features on a different evaluation set}
\label{app:fea_eval}
\begin{figure*}
    \centering
    \includegraphics[width=\textwidth]{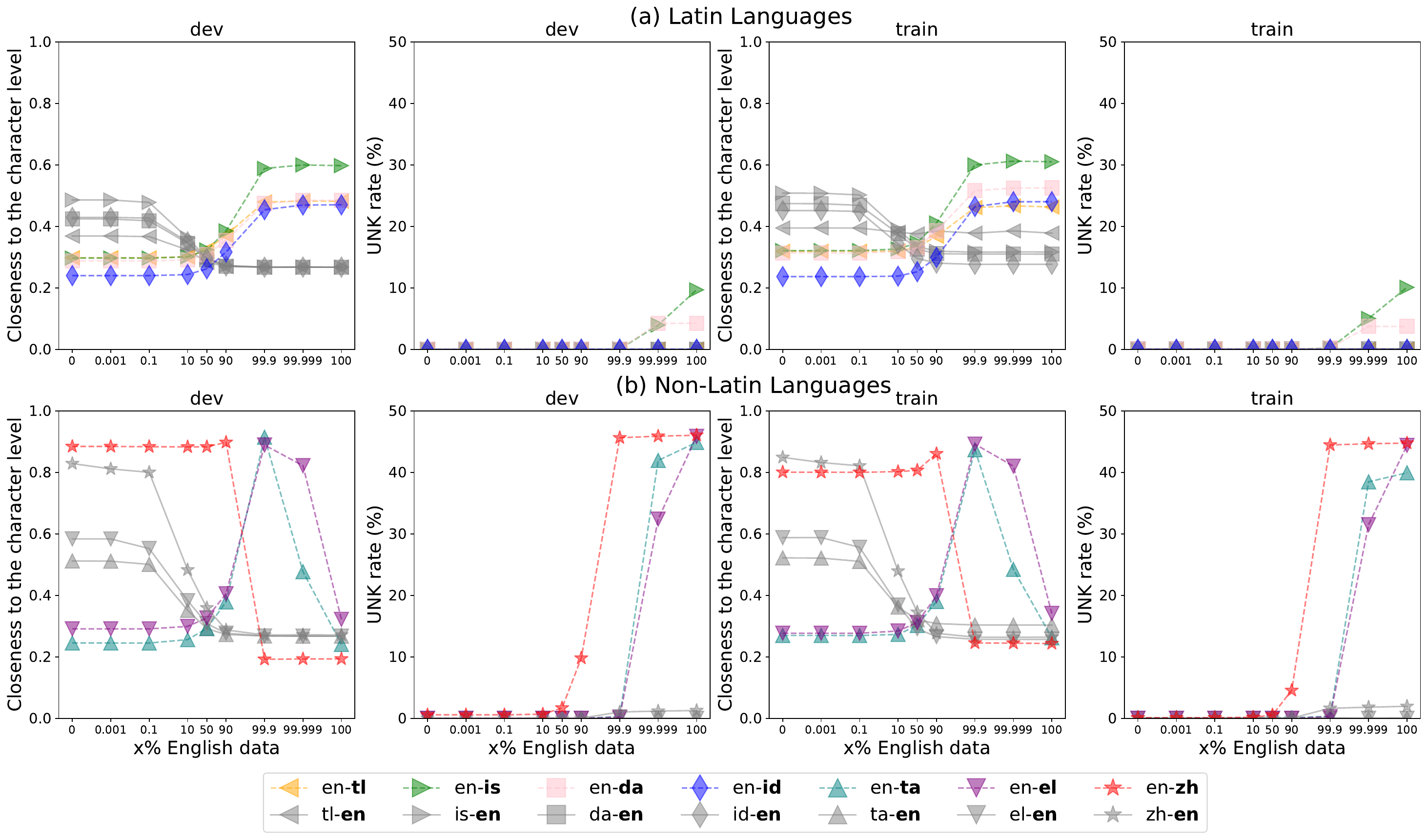}
    \vspace{-10pt}
    \caption{In each row, the first two subplots are features computed on the \textsc{Flores}101 dev set; the second two subplots are features computed on a subset of our training set. Markers share the same meanings as Figure~\ref{fig:main_res}. X axes are in log10 scale.}
    \vspace{-7pt}
    \label{fig:dev_train}
\end{figure*}

In the main paper, we compute intermediate features on \textsc{Flores}101 devtest set where we also report translation performances. However, usually, we are blind to the testing sets. We want to ask whether the same pattern can still be observed when we get the features on a different evaluation set. Therefore, we get features from the dev set and a subset of the training set (with 5000 sentence pairs). The first and the second two columns of Figure~\ref{fig:dev_train} illustrate the features obtained from the dev and training set respectively. Compared to the features in Figure~\ref{fig:main_res}, very similar trends are observed, except for slightly different thresholds. When evaluating on the dev set, English to Icelandic (en-is) and English to Danish (en-da) get worse when Icelandic and Danish have 4.0\% and 4.2\% UNK rates respectively; English to Chinese (en-zh) drops when Chinese UNK rate is 9.8\%; English to Tamil (en-ta) and English to Greek (en-el) drops when the closeness to the character level is 0.91 and 0.89 respectively. On the subset of the training set, when performances deteriorate, Icelandic, Danish, and Chinese have 5.0\%, 3.7\%, and 4.6\% UNK rates respectively, and Tamil and Greek have 0.87 and 0.89 closeness to the character level respectively. Overall, despite the thresholds being lower (3.7\% UNK rate and 0.87 closeness to the character level), the same takeaways still hold when getting features from different evaluation sets.

\subsection{Bilingual Ablations}

\subsubsection{Smaller Model}
\label{app:smaller}
\begin{figure*}
    \centering
    \includegraphics[width=0.7\textwidth]{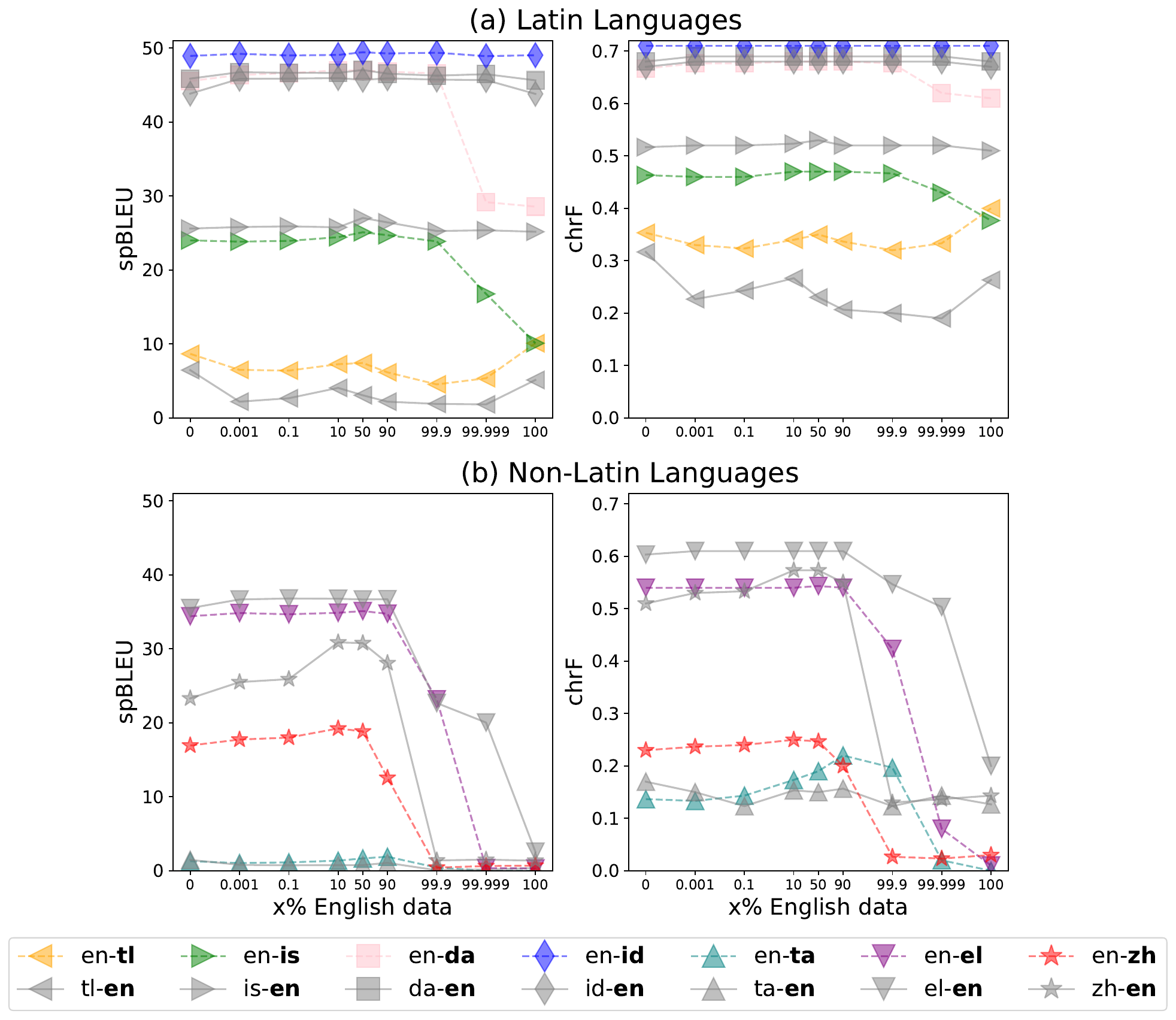}
    \vspace{-5pt}
    \caption{Translation results of bilingual experiments with a smaller model (Transformer 6-6). Markers share the same meanings as Figure~\ref{fig:main_res}. X axes are in log10 scale.}
    \vspace{-10pt}
    \label{fig:smaller_model}
\end{figure*}

The translation results of using a smaller model (Transformer 6-6) are shown in Figure~\ref{fig:smaller_model}. We observe that performances drop at the same English percentages as Figure~\ref{fig:main_res}. Meanwhile, the features are the same as Figure~\ref{fig:main_res}. Thus, the exact same conclusions are obtained. 

\begin{table}
\begin{center}
\resizebox{0.4\textwidth}{!}{%
\begin{tabular}{l|ccc}
\toprule 
& \textbf{100\%} & \textbf{100\%+char} & \textbf{best} \\
\midrule
en-ta & 0.0 & 0.1 & \bf1.9 \\
ta-en & 0.2 & 0.1 &  \bf1.1\\
\midrule
en-el & 0.3 & 18.6 &  \bf35.1\\
el-en & 2.7 & 18.5 &  \bf36.7 \\
\midrule
en-zh & 0.6 &  \bf20.0 & 19.2\\
zh-en & 1.5 &  \bf31.2 & 30.9\\
\toprule
\end{tabular}
}
\end{center}
\vspace{-10pt}
\caption{Translation results (spBLEU scores) of adding the non-Latin language's characters to the vocabulary at English=100\% (\textbf{100\%+char}). For comparison, the \textbf{100\%} column shows the results before adding characters and the \textbf{best} column shows the best results out of all percentages. }
\vspace{-15pt}
\label{tab:add_char}
\end{table}

\subsubsection{BPE}
\label{app:bpe}
\begin{figure*}
    \centering
    \includegraphics[width=\textwidth]{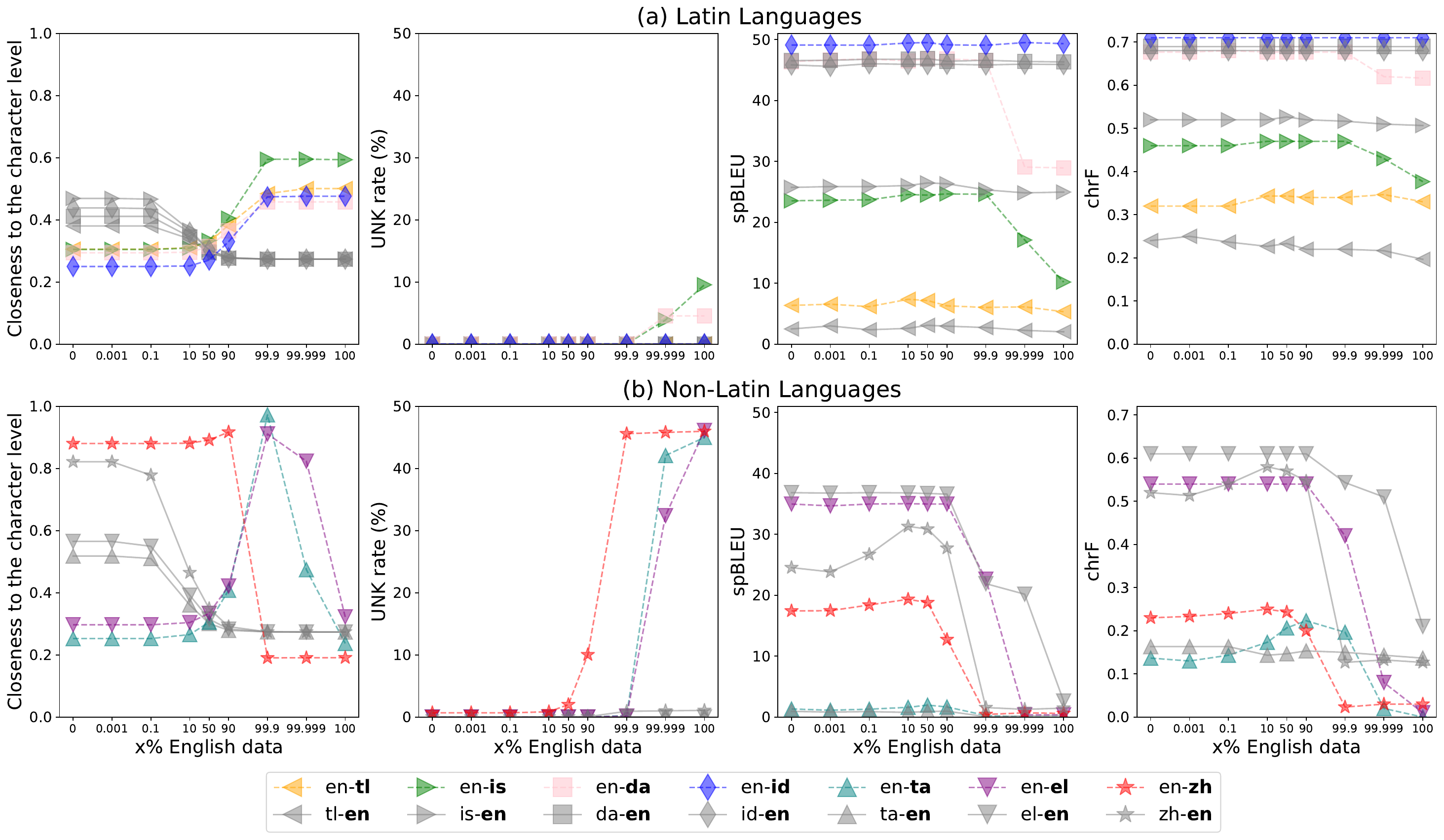}
    \vspace{-18pt}
    \caption{Intermediate features and translation results of bilingual experiments with a BPE tokenizer. Markers share the same meanings as Figure~\ref{fig:main_res}. X axes are in log10 scale.}
    \vspace{-10pt}
    \label{fig:bpe}
\end{figure*}

The features and translation results of using a BPE tokenizer are shown in Figure~\ref{fig:bpe}. It shares the same trends with Figure~\ref{fig:main_res} but with slightly higher thresholds: English to Icelandic (en-is) and English to Danish (en-da) deteriorate when Icelandic and Danish have 3.9\% and 4.6\% UNK rates respectively; English to Chinese (en-zh) drops when Chinese UNK rate is 10.0\%; English to Tamil (en-ta) and English to Greek (en-el) get worse when the closeness to the character level is 0.97 and 0.91 respectively. 

\subsubsection{Larger Vocabulary}
\label{app:larger}
\begin{figure*}
    \centering
    \includegraphics[width=\textwidth]{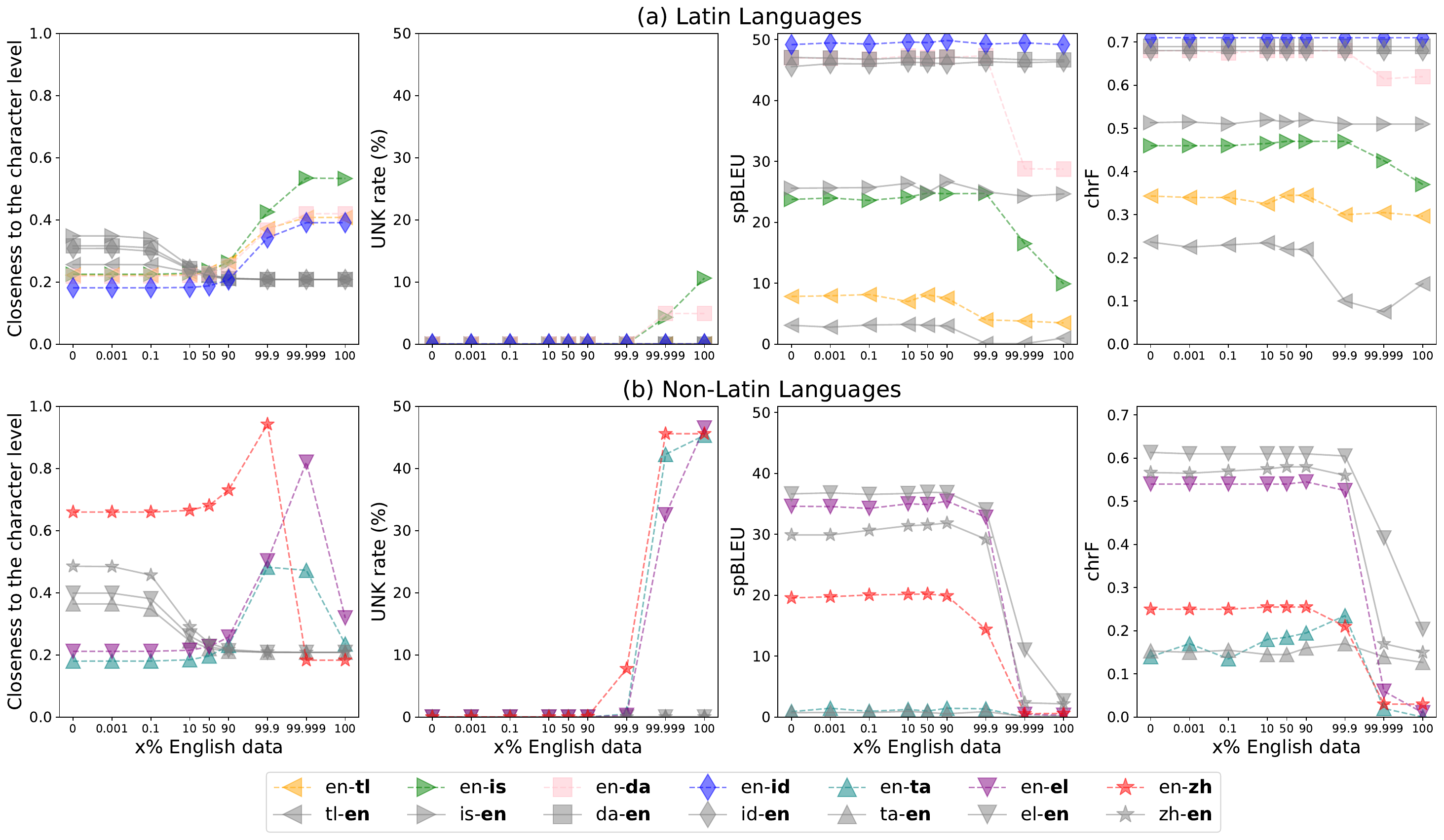}
    \vspace{-18pt}
    \caption{Intermediate features and translation results of bilingual experiments with a 32K vocabulary. Markers share the same meanings as Figure~\ref{fig:main_res}. X axes are in log10 scale.}
    \vspace{-10pt}
    \label{fig:larger_vocab}
\end{figure*}

The features and translation results of using a 32K vocabulary are shown in Figure~\ref{fig:larger_vocab}. It has two distinctions from Figure~\ref{fig:main_res} which are discussed in Section~\ref{sec:bi-ablation}.

\subsubsection{Byte-fallback}
\label{app:byte}
\begin{figure*}
    \centering
    \includegraphics[width=\textwidth]{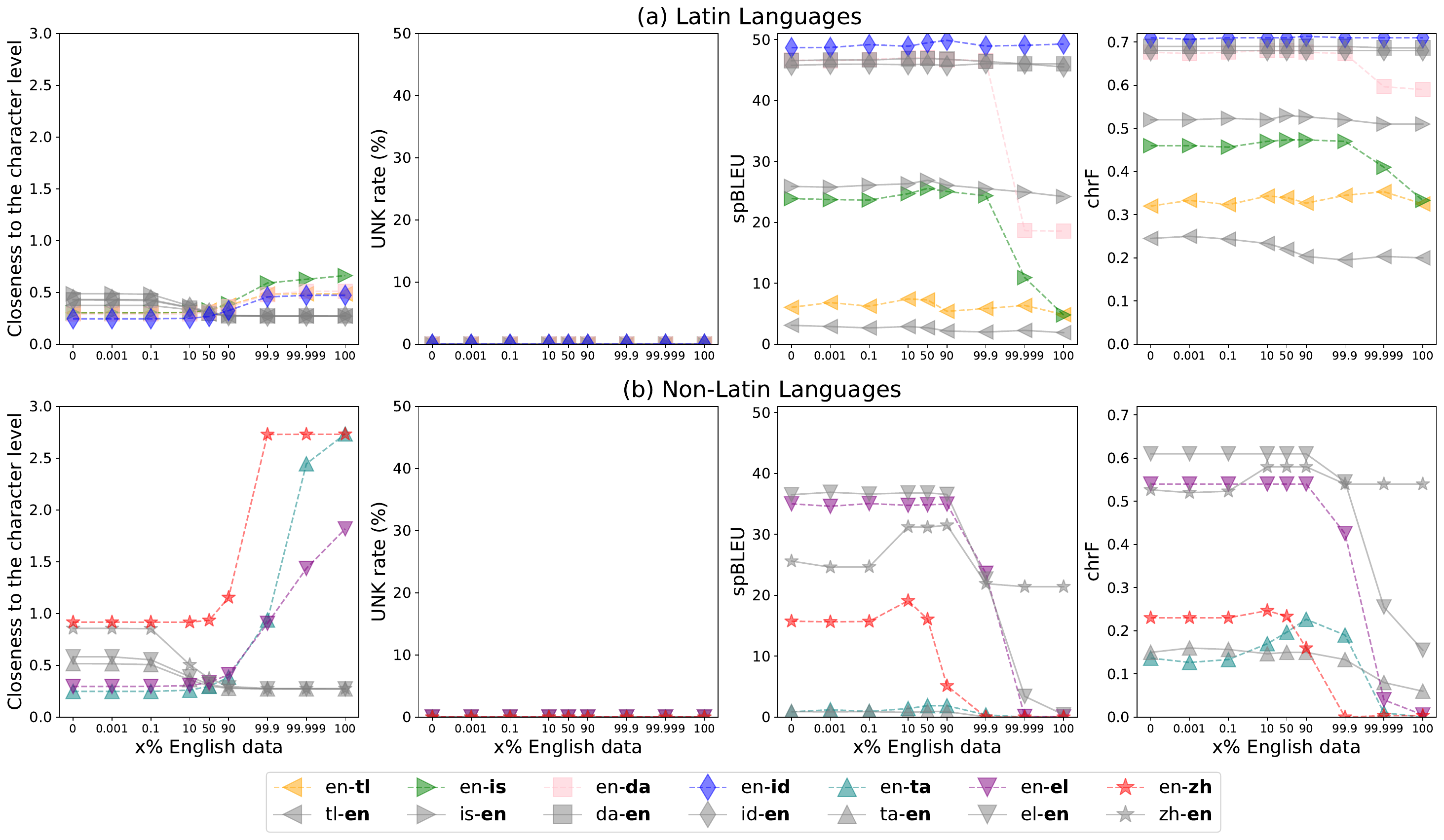}
    \vspace{-18pt}
    \caption{Intermediate features and translation results of bilingual experiments with byte-fallback. Note that here the UNK rates are all 0, and closeness to the character level can be larger than 1 because one character can be represented by multiple bytes. Markers share the same meanings as Figure~\ref{fig:main_res}. X axes are in log10 scale.}
    \vspace{-10pt}
    \label{fig:byte}
\end{figure*}

The features and translation results of using a 32K vocabulary are shown in Figure~\ref{fig:byte}. Discussions are in Section~\ref{sec:bi-ablation}.

\subsubsection{Adding characters}
\label{app:adding_char}
Table~\ref{tab:add_char} shows the results of adding the non-Latin language's characters to the vocabulary when English=100\%. 

\subsection{Multilingual Results and Ablations}
\subsubsection{Main Translation Results (chrF) and Features}
\label{app:multi_main}
\begin{figure*}
    \centering
    \includegraphics[width=\textwidth]{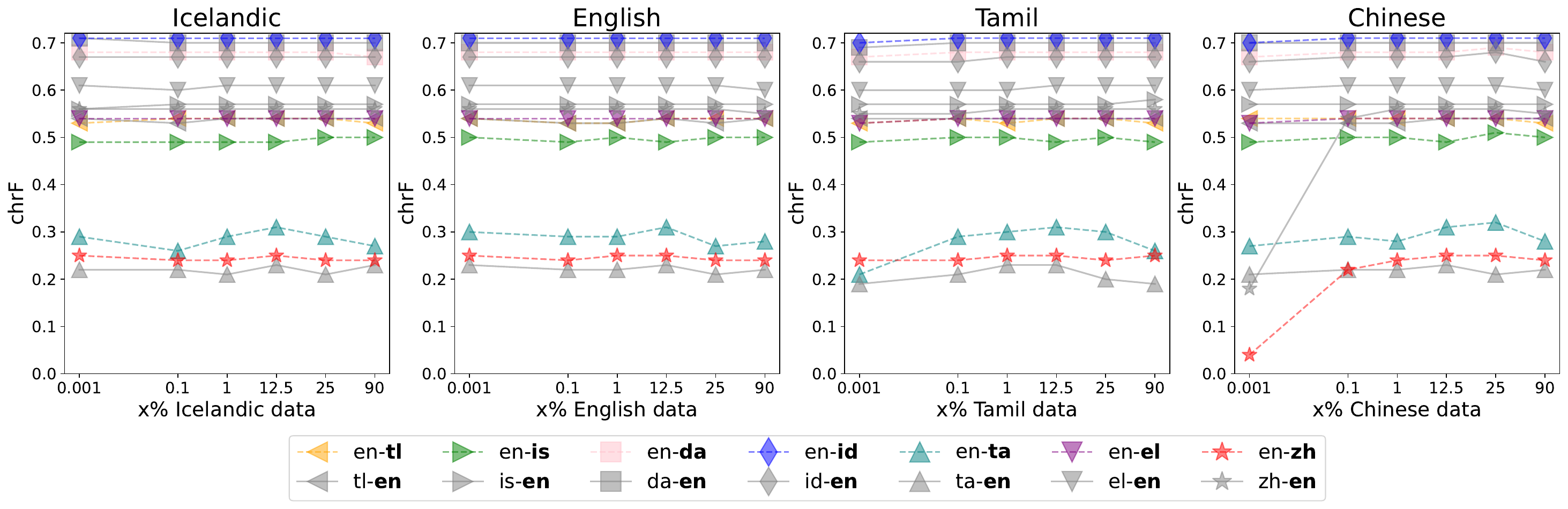}
    \vspace{-18pt}
    \caption{Translation results (chrF) of our main multilingual experiments. Markers have the same meanings as Figure~\ref{fig:multi-spbleu}. X axes are in log10 scale.}
    \vspace{-5pt}
    \label{fig:multi-chrf}
\end{figure*}

\begin{figure*}
    \centering
    \includegraphics[width=\textwidth]{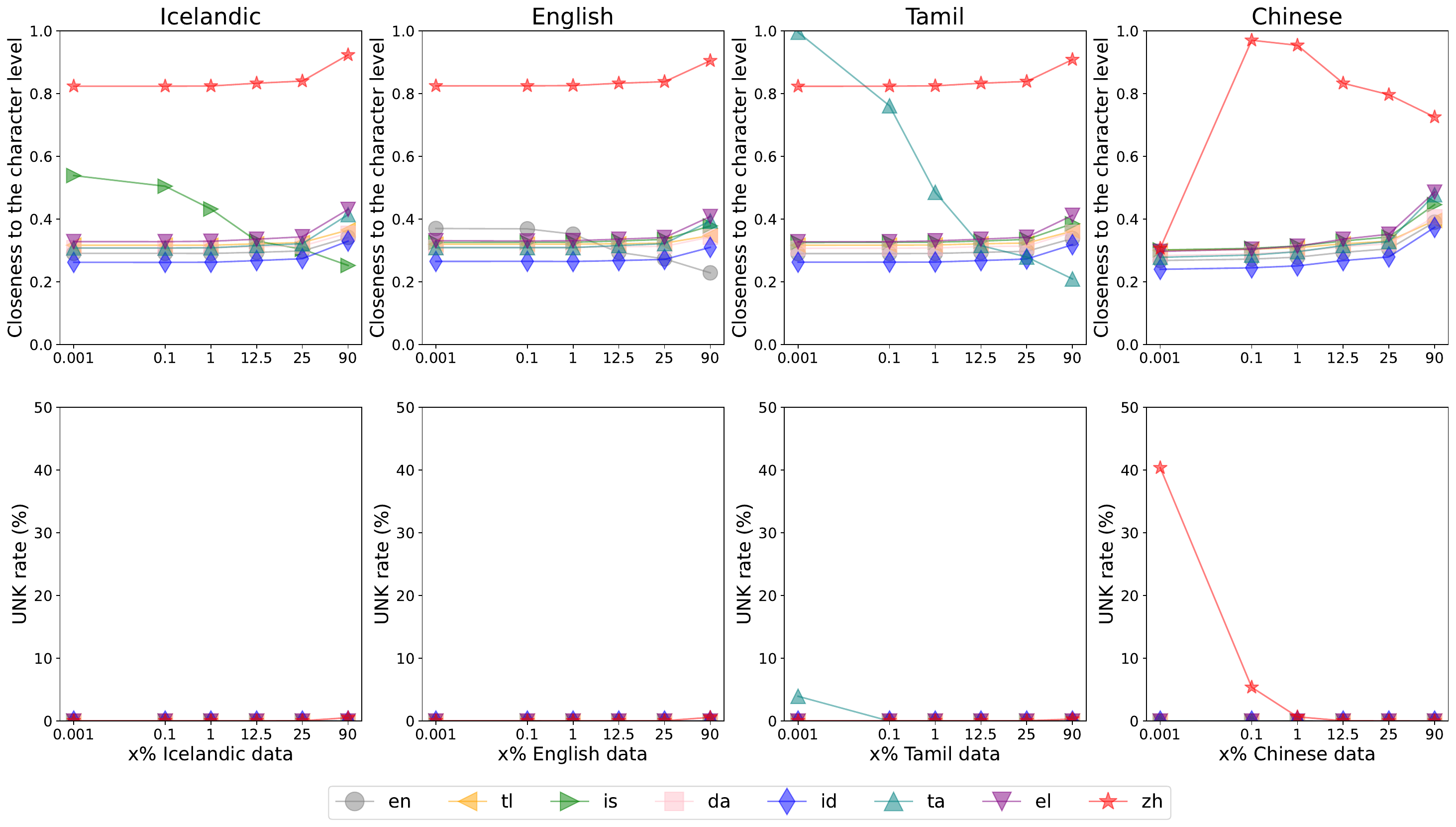}
    \vspace{-18pt}
    \caption{Intermediate features of our main multilingual experiments. Different from Figure~\ref{fig:multi-spbleu}, here, marker shapes and colors both denote the language. E.g., {\color{teal}-}$\begingroup\color{teal}\blacktriangle\endgroup${\color{teal}-} (ta) denotes Tamil features. X axes are in log10 scale.}
    \vspace{-5pt}
    \label{fig:multi-feature}
\end{figure*}

Figure~\ref{fig:multi-chrf} shows the chrF scores of our main multilingual experiments. It shares the same trends with Figure~\ref{fig:multi-spbleu}. Figure~\ref{fig:multi-feature} show the features of each of the 8 languages. Different from features in bilingual experiments, here, we do not have to distinguish language pairs because all languages are mixed together to train one joint vocabulary. 

\subsubsection{Byte-fallback}
\label{app:multi_byte}

\begin{figure*}
    \centering
    \includegraphics[width=\textwidth]{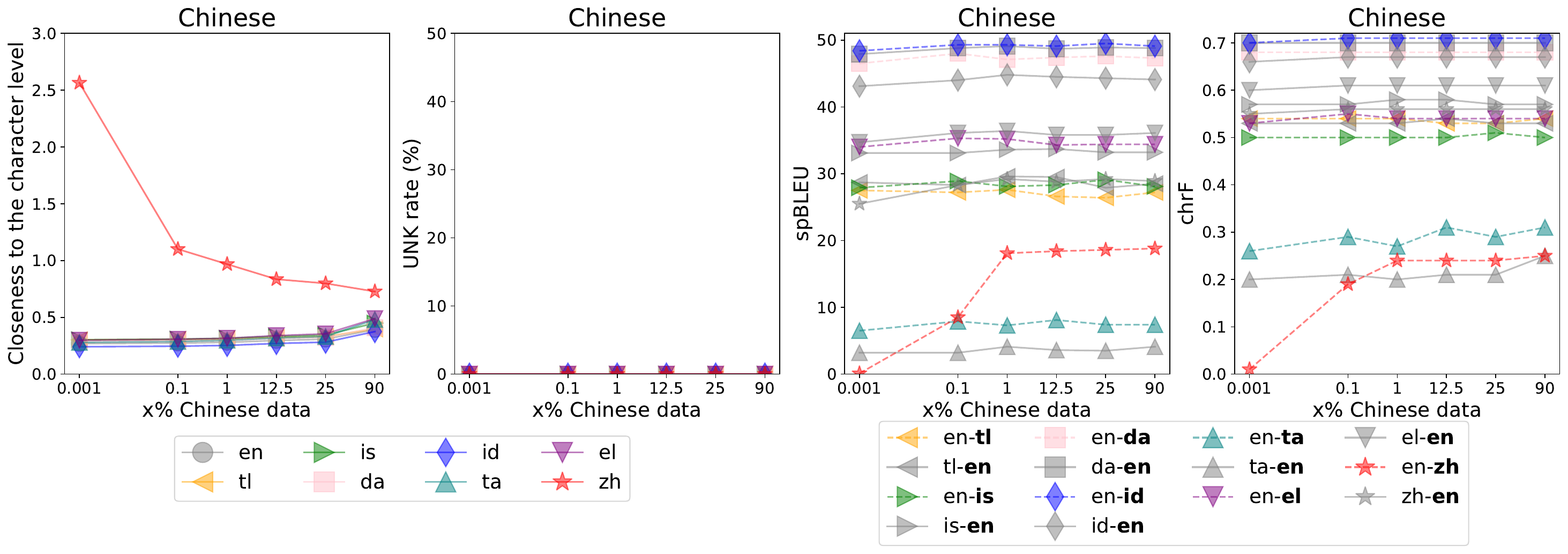}
    \vspace{-18pt}
    \caption{Intermediate features and translation results of the multilingual experiments with byte-fallback. Markers of the first two subplots have the same meanings as Figure~\ref{fig:multi-feature}, and markers of the second two subplots have the same meanings as Figure~\ref{fig:multi-spbleu}. X axes are in log10 scale.}
    \vspace{-5pt}
    \label{fig:multi-zh-byte}
\end{figure*}

Figure~\ref{fig:multi-zh-byte} illustrates the translation results and features of the multilingual experiments with byte-fallback when only the Chinese percentage varies. Discussions are in Section~\ref{sec:multi-results}.

\subsection{Examples}
\label{app:examples}
Table~\ref{tab:examples} are examples of how sentences in English, Indonesian, and Chinese are tokenized at different English percentages under the main bilingual setting (Section~\ref{sec:bi-exp}). 

\begin{table*}
\small
\begin{center}
\begin{tabular}{l|p{0.42\textwidth}|p{0.42\textwidth}}
\toprule 
x\% English & English & Indonesian \\
\midrule
0 & \_" W e \_no w \_ha ve \_4 - mon th - ol d \_mi ce \_th at \_a re \_non - dia be tic \_th at \_ us ed \_to \_be \_dia be tic ," \_he \_ad de d . & \_“ S a at \_ini \_ada \_men ci t \_umur \_4 \_bulan \_non dia bet es \_yang \_dulu nya \_diabetes ,” \_tambah nya .\\
\midrule
0.1 & \_" W e \_no w \_ha ve \_4 - mon th - ol d \_mi ce \_th at \_a re \_non - dia be tic \_th at \_ us ed \_to \_be \_dia be tic ," \_he \_ad de d . & \_“ S a at \_ini \_ada \_men ci t \_umur \_4 \_bulan \_non dia bet es \_yang \_dulu nya \_diabetes ,” \_tambah nya .\\
\midrule
50 & \_" We \_now \_have \_4 - mon th - old \_mi ce \_that \_are \_non - dia be tic \_that \_used \_to \_be \_dia be tic ," \_he \_added . & \_“ S a at \_ini \_ada \_men ci t \_umur \_4 \_bulan \_non dia bet es \_yang \_dulu nya \_diabetes ,” \_tambah nya .\\
\midrule
99.9 & \_" We \_now \_have \_4 - mon th - old \_mi ce \_that \_are \_non - d ia be tic \_that \_used \_to \_be \_di a be tic ," \_he \_added . & \_“ S a at \_in i \_a da \_men ci t \_ um ur \_4 \_bu lan \_non di ab et es \_ya ng \_du lu nya \_diabetes ,” \_ta mb ah nya .\\
\midrule
100 & \_" We \_now \_have \_4 - mon th - old \_mi ce \_that \_are \_non - d ia be tic \_that \_used \_to \_be \_di a be tic ," \_he \_added . & \_“ S a at \_in i \_a da \_men ci t \_ um ur \_4 \_b ul an \_non di ab et es \_ya ng \_du lu ny a \_diabetes ,” \_ta mb ah ny a .\\
\midrule
\midrule
x\% English & English & Chinese \\
\midrule
0 & \_ " W e \_ n o w \_ h a v e \_ 4 - m on t h - o l d \_ m ic e \_ t h at \_ ar e \_ n on - d i a b et ic \_ t h at \_ u s e d \_ t o \_ b e \_ d i a b et ic , " \_ h e \_ ad d e d . & \begin{CJK}{UTF8}{gbsn}\_ 他 补 充 道 :“ 我们 现在 有 \_ 4 \_ 个月 大 没有 糖 尿 病 的 老 鼠 , 但 它们 曾 经 得 过 该 病 。”\end{CJK}\\
\midrule
0.1 & \_ " W e \_ n o w \_ h a v e \_ 4 - m on th - o l d \_ m ic e \_ th at \_ ar e \_ n on - d i a b et ic \_ th at \_ u s ed \_ t o \_ b e \_ d i a b et ic , " \_ h e \_ ad d ed . & \begin{CJK}{UTF8}{gbsn}\_ 他 补 充 道 :“ 我们 现在 有 \_ 4 \_ 个月 大 没有 糖 尿 病 的 老 鼠 , 但 它们 曾 经 得 过 该 病 。”\end{CJK} \\
\midrule
50 &  \_" We \_now \_have \_4 - mon th - old \_mi ce \_that \_are \_no n - dia be tic \_that \_used \_to \_be \_ dia be tic , " \_he \_add ed . & \begin{CJK}{UTF8}{gbsn}\_ 他 补 充 道 :“ 我们 现在 有 \_4 \_ 个 月 大 没有 糖 尿 病 的 老 鼠 , 但 它 们 曾 经 得 过 该 病 。”\end{CJK}\\
\midrule
99.9 & \_" We \_now \_have \_4 - mon th - old \_ m ice \_that \_are \_non - dia be tic \_that \_used \_to \_be \_ dia be tic ," \_he \_added . & \_ <unk> : “ <unk> \_4 \_ <unk> , <unk> ”\\
\midrule
100 & \_" We \_now \_have \_ 4 - mon th - old \_ m ice \_that \_are \_non - dia be tic \_that \_used \_to \_be \_ dia be tic ," \_he \_added . & \_ <unk> : “ <unk> \_ 4 \_ <unk> , <unk> ”\\
\toprule
\end{tabular}
\end{center}
\vspace{-13pt}
\caption{Examples of how sentences in English, Indonesian, and Chinese are tokenized at different English percentages under our main bilingual setting (Section~\ref{sec:bi-exp}). The sentence is the first sentence of \textsc{Flores}101 devtest set. Subwords are separated by whitespaces, and unknown tokens are replaced by `<unk>'. }
\vspace{-12pt}
\label{tab:examples}
\end{table*}

\end{document}